\documentclass{article} 
\usepackage{nips13submit_e,times}
\usepackage{hyperref}
\usepackage{url}
\usepackage{graphicx}
\usepackage{amsmath}
\usepackage{amsfonts}
\usepackage{fancyhdr}
\usepackage[top=3cm, bottom=4cm, left=3.5cm, right=3.5cm, headheight=20pt, headsep=\baselineskip, includehead]{geometry}

\usepackage{float}
\usepackage{subcaption}
\floatstyle{ruled}
\newfloat{pseudocode}{ht}{ext}
\floatname{pseudocode}{Algorithm}
\usepackage[round,comma]{natbib}
\DeclareMathOperator*{\argmax}{argmax}

\title{Double Q($\sigma$) and Q($\sigma, \lambda$) \\ Unifying Reinforcement Learning Control Algorithms}

\nipsfinalcopy 

\begin{document}

\thispagestyle{empty}
\maketitle

\begin{abstract}
Temporal-difference (TD) learning is an important field in reinforcement learning. Sarsa and Q-Learning are among the most used TD algorithms. The Q($\sigma$) algorithm (\cite{sutton2017}) unifies both.
This paper extends the Q($\sigma$) algorithm to an on-line multi-step algorithm Q($\sigma, \lambda$) using eligibility traces and introduces Double Q($\sigma$) as the extension of Q($\sigma$) to double learning. 
Experiments suggest that the new Q($\sigma, \lambda$) algorithm can outperform the classical TD control methods Sarsa($\lambda$), Q($\lambda$) and Q($\sigma$).
\end{abstract}

\section{Introduction}

Reinforcement Learning is a field of machine learning addressing the problem of sequential decision making. It is formulated as an interaction of an agent and an environment over a number of discrete time steps $t$. At each time step the agent chooses an action $A_t$ based on the environment's state $S_t$. The environment takes $A_t$ as an input and returns the next state observation $S_{t+1}$ and reward $R_{t+1}$, a scalar numeric feedback signal.

The agent is thereby following a policy $\pi$, which is the behavior function mapping a state to action probabilities

\begin{equation} 
\pi(a | s) = P(A_t = a | S_t = s).
\end{equation}


The agent's goal is to maximize the return $G_t$ which is the sum of discounted rewards,

\begin{equation} 
G_t = R_{t+1} + \gamma R_{t+2} + \gamma^2 R_{t+3} + ... = \sum_{k=0}^{T-1} \gamma^k R_{t+1+k},
\end{equation}

where $\gamma \in [0, 1]$ is the discount factor and $T$ is the length of the episode or infinity for a continuing task.

While rewards are short-term signals about the goodness of an action, values represent the long-term value of a state or state-action pair. The action value function $q_\pi(s, a)$ is defined as the expected return taking action $a$ from state $s$ and thereafter following policy $\pi$:

\begin{equation} 
q_\pi(s, a) = \mathbb{E}_\pi[G_t | S_t = s, A_t = a].
\label{actionvalue}
\end{equation}

Value-based reinforcement learning is concerned with finding the optimal action value function $q_* = \max_\pi q_\pi$. Temporal-difference learning is a class of model-free methods which estimates $q_\pi$ from sample transitions and iteratively updates the estimated values using observed rewards and estimated values of successor actions. At each step an update of the following form is applied:

\begin{equation} 
Q(S_t, A_t) \leftarrow Q(S_t, A_t) + \alpha \; \delta_t,
\label{equ:update}
\end{equation}

where $Q$ is an estimate of $q_\pi$, $\alpha$ is the step size and $\delta_t$ is the TD error, the difference between our current estimate and a newly computed target value. The following TD control algorithms can all be characterized by their different TD errors.

When the action values $Q$ are represented as a table we call this tabular reinforcement learning, else we speak of approximate reinforcement learning, e.g. when using a neural network to compute the action values. For sake of simplicity the following analysis is done for tabular reinforcement learning but can be easily extended to function approximation.

\section[TD control algorithms]{TD control algorithms: From Sarsa to Q($\sigma$)}

Sarsa (\cite{sarsa}) is a temporal-difference learning algorithm which samples states and actions using an $\epsilon$-greedy policy and then updates the $Q$ values using Equation \ref{equ:update} with the following TD error 

\begin{equation} 
\delta_t = R_{t+1} + \gamma Q(S_{t+1}, A_{t+1}) - Q(S_t, A_t).
\label{sarsa}
\end{equation}

The term $R_{t+1} + \gamma Q(S_{t+1}, A_{t+1})$ is called the TD target and consists of the reward plus the discounted value of the next state and next action.

Sarsa is an on-policy method, i.e. the TD target consists of $Q(S_{t+1}, A_{t+1})$, where $A_{t+1}$ is sampled using the current policy. In general the policy used to sample the state and actions - the so called behaviour-policy $\mu$ - can be different from the target policy $\pi$, which is used to compute the TD target. If behaviour and target policy are different we call this off-policy learning. An example for an off-policy TD control algorithm is the well known Q-Learning algorithm proposed by \cite{watkins1989}. As in Sarsa states and actions are sampled using an exploratory behaviour policy, e.g. an $\epsilon$-greedy policy, but the TD target is computed using the greedy policy with respect to the current Q values. The TD error of Q-Learning is

\begin{equation} 
\delta_t = R_{t+1} + \gamma \max_{a'} Q(S_{t+1}, a') - Q(S_t, A_t).
\label{qlearning}
\end{equation}

Expected Sarsa generalizes Q-Learning to arbitrary target policies. The TD error is

\begin{equation} 
\delta_t = R_{t+1} + \gamma \sum_{a'} \pi(a' | S_{t+1}) Q(S_{t+1}, a') - Q(S_t, A_t).
\label{equ:expectedsarsa}
\end{equation}

The current state-action pair is updated using the expectation of all subsequent action values with respect to the action value. Q-Learning is a special case of Expected Sarsa if $\pi$ is the greedy policy with respect to $Q$ (\cite{sutton2017}). Of course Expected Sarsa could also be used as an on-policy algorithm if the target policy is chosen to be the same as the behaviour policy (\cite{expsarsa}).

\cite{sutton2017} propose a new TD control algorithm called Q($\sigma$) which unifies Sarsa and Expected Sarsa. The TD target of this new algorithm is a weighted mean of the Sarsa and Expected Sarsa TD targets, where the parameter $\sigma$ controls the weighting. Q(1) is equal to Sarsa and Q(0) is equal to Expected Sarsa. For intermediate values of $\sigma$ new algorithms are obtained, which can achieve better performance (\cite{deasis2017}).

The TD error of Q($\sigma$) is 

\begin{equation} 
\delta_t = R_{t+1} + \gamma ( \sigma Q(S_{t+1}, A_{t+1}) + (1 - \sigma) \sum_{a'} \pi(a' | S_{t+1}) Q(S_{t+1}, a') ) - Q(S_t, A_t).
\end{equation}

\section{Q($\sigma, \lambda$): An on-line multi-step algorithm}

The TD methods presented so far are one-step methods, which use only rewards and values from the next step $t+1$. These can be extended to use eligibility traces to incorporate data of multiple time steps.

An eligibility trace is a scalar numeric value for each state-action pair. Whenever a state-action pair is visited its eligibility is increased, if not, the eligibility fades away over time. State-action pairs visited often will have a higher eligibility than those visited less frequently and state-action pairs visited recently will have a higher eligibility than those visited long time ago.

The accumulating eligibility trace (\cite{eligibility}) uses an update of the form

\begin{equation}
    E_{t+1}(s, a) =
\begin{cases}
    \gamma \lambda E_t(s, a) + 1, & \text{if } A_t = a, S_t = s\\
    \gamma \lambda E_t(s, a) ,    & \text{otherwise.}
\end{cases}
\end{equation}

Whenever taking action $A_t$ in state $S_t$ the eligibility of this pair is increased by 1 and for all states and actions decreased by a factor $\gamma \lambda$, where $\lambda$ is the trace decay parameter. 

Then all state-action pairs are updated according to their eligibility trace

\begin{equation}
Q(s, a) \leftarrow Q(s, a) + \alpha \delta_t \, E_t(s, a) 
\end{equation}

The corresponding algorithm using the one-step Sarsa TD error and an update using eligibility traces is called Sarsa($\lambda$). Though it looks like a one-step algorithm, it is in fact a multi-step algorithm, because the current TD error is assigned back to all previously visited states and actions weighted by their eligibility. 

For off-policy algorithms like Q-Learning different eligibility updates have been proposed. Watkin's Q($\lambda$) uses the same updates as long as the greedy action is chosen by the behaviour policy, but sets the $Q$ values to 0, whenever a non-greedy action is chosen assigning credit only to state-action pairs we would actually have visited if following the target policy $\pi$ and not the behaviour policy $\mu$. More generally the eligibility is weighted by the target policy's probability of the next action. The update rule is then

\begin{equation}
    E_{t+1}(s, a) =
\begin{cases}
    \gamma \lambda E_t(s, a) \pi(A_{t+1} | S_{t+1}) + 1, & \text{if } A_t = a, S_t = s\\
    \gamma \lambda E_t(s, a) \pi(A_{t+1} | S_{t+1}),    & \text{otherwise.}
\end{cases}
\end{equation}

Whenever an action occurs, which is unlikely in the target policy, the eligibility of all previous states is decreased sharply. If the target policy is the greedy policy, the eligibility will be set to 0 for the complete history.

In this paper we introduce a new kind of eligibility trace update to extend the Q($\sigma$) algorithm to an on-line multi-step algorithm, which we will call Q($\sigma, \lambda$).
Recall that the one-step target of Q($\sigma$) is a weighted average between the on-policy Sarsa and off-policy Expected Sarsa targets weighted by the factor $\sigma$:

\begin{equation} 
\delta_t = R_{t+1} + \gamma ( \sigma Q(S_{t+1}, A_{t+1}) +  (1 - \sigma) \sum_{a'} \pi(a' | S_{t+1}) Q(S_{t+1}, a') ) - Q(S_t, A_t)
\end{equation}

In this paper we propose to weight the eligibility accordingly with the same factor $\sigma$. The eligibility is then a weighted average between the on-policy eligibility used in Sarsa($\lambda$) and the off-policy eligibility used in $Q(\lambda)$. The eligibility trace is updated at each step by

\begin{equation}
    E_{t+1}(s, a) =
\begin{cases}
    \gamma \lambda E_t(s, a) (\sigma + (1 - \sigma) \pi(A_{t+1} | S_{t+1})) + 1, & \text{if } A_t = a, S_t = s\\
    \gamma \lambda E_t(s, a) (\sigma + (1 - \sigma) \pi(A_{t+1} | S_{t+1})),    & \text{otherwise.}
\end{cases}
\end{equation}

When $\sigma = 0$ the one-step target of Q($\sigma$) is equal to the Sarsa one-step target and therefore the eligibility update reduces to the standard accumulate eligibility trace update. When $\sigma = 1$ the one-step target of Q($\sigma$) is equal to the Expected Sarsa target and accordingly the eligibility is weighted by the target policy's probability of the current action. For intermediate values of $\sigma$ the eligibility is weighted in the same way as the TD target. \cite{deasis2017} showed that n-step Q($\sigma$) with an intermediate or dynamic value of $\sigma$ can outperform Q-Learning and Sarsa. By extending this algorithm to an on-line multi-step algorithm we can make use of the good initial performance of Sarsa($\lambda$) combined with the good asymptotic performance of Q($\lambda$). In comparison to the n-step Q($\sigma$) algorithm (\cite{deasis2017}) the new Q($\sigma$, $\lambda$) algorithm can learn on-line and is therefore likely to learn faster. 

Pseudocode for tabular episodic Q($\sigma, \lambda$) is given in Algorithm \ref{algorithm1}. This can be easily extended to continuing tasks and to function approximation using one eligibility per weight of the function approximator.

\begin{pseudocode}
\begin{itemize}
  \item[] Initialize $Q(s, a) \quad \forall s \in \mathcal{S}, a \in \mathcal{A}$ 
  \item[] Repeat for each episode:
  \begin{itemize}
  \item[] $E(s, a) \leftarrow 0 \quad \forall s \in \mathcal{S}, a \in \mathcal{A}$
  \item[] Initialize $S_0 \neq$ terminal
  \item[] Choose $A_0$, e.g. $\epsilon$-greedy from $Q(S_0, .)$
  \item[] Loop for each step of episode:
  \begin{itemize}
    \item[] Take action $A_t$, observe reward $R_{t+1}$ and next state $S_{t+1}$
    \item[] Choose next action $A_{t+1}$, e.g. $\epsilon$-greedy from $Q(S_{t+1}, .)$
    \item[] $\delta = R_{t+1} + \gamma \, (\sigma \, Q(S_{t+1}, A_{t+1}) + (1- \sigma) \; \sum_{a'} \,\pi(a'|S_{t+1}) \; Q(S_{t+1}, a')) - Q(S_{t}, A_t) $
    \item[] $E(S_t, A_t) \leftarrow E(S_t, A_t) + 1$
    \item[] $Q(s, a) \leftarrow Q(s, a) + \alpha \, \delta \, E(s, a) \quad \forall s \in \mathcal{S}, a \in \mathcal{A}$
    \item[] $E(s, a) \leftarrow \gamma \lambda E(s, a) ( \sigma + (1 - \sigma) \pi(A_{t+1} | S_{t+1}) ) \quad \forall s \in \mathcal{S}, a \in \mathcal{A}$
    \item[] $A_t \leftarrow A_{t+1}$, $S_t \leftarrow S_{t+1}$
    \item[] If $S_t$ is terminal: Break
  \end{itemize}
\end{itemize}
\end{itemize}
\caption{Q($\sigma, \lambda$)}
\label{algorithm1}
\end{pseudocode}

\section{Double Q($\sigma$) Algorithm}

Double learning is another extension of the basic algorithms. It has been mostly studied with Q-Learning \cite{hasselt2010} and prevents the overestimation of action values when using Q-Learning in stochastic environments. The idea is to use decouple action selection (which action is the best one?) and action evaluation (what is the value of this action?). The implementation is simple, instead of using only one value function we will use two value functions $Q_A$ and $Q_B$. Actions are sampled due to an $\epsilon$-greedy policy with respect to $Q_A + Q_B$. Then at each step either $Q_A$ or $Q_B$ is updated, e.g. if $Q_A$ is selected by

\begin{align}
    Q_A(S_t, A_t) &\leftarrow Q_A(S_t, A_t) + \alpha (R_{t+1} + \gamma Q_B(\argmax_{a \in \mathcal{A}} Q_A(S_{t+1}, a)) - Q_A(S_t, A_t)) \\
    Q_B(S_t, A_t) &\leftarrow Q_B(S_t, A_t) + \alpha (R_{t+1} + \gamma Q_A(\argmax_{a \in \mathcal{A}} Q_B(S_{t+1}, a)) - Q_B(S_t, A_t))
\end{align}

Double learning can also be used with Sarsa and Expected Sarsa as proposed by \cite{ganger2016}. Using double learning these algorithms can be more robust and perform better in stochastic environments. The decoupling of action selection and action evaluation is weaker than in Double Q-Learning because the next action $A_{t+1}$ is selected according to an $\epsilon$-greedy behavior policy using $Q_A + Q_B$ and evaluated either with $Q_A$ or $Q_B$. For Expected Sarsa the policy used for the target in Equation \ref{equ:expectedsarsa} could be the $\epsilon$-greedy behavior policy as proposed by \cite{ganger2016}, but it is probably better to use a policy according to $Q_A$ (if updating $Q_A$), because then it can also be used off-policy with Double Q-Learning as a special case, if $\pi$ is the greedy policy with respect to $Q_A$.

In this paper we propose the extension of double learning to Q($\sigma$) - Double Q($\sigma$) - to obtain a new algorithm with the good learning properties of double learning, which generalizes (Double) Q-Learning, (Double) Expected Sarsa and (Double) Sarsa. Of course Double Q($\sigma$) can also be used with eligibility traces.

Double Q($\sigma$) has the following TD error when $Q_A$ is selected,

\begin{equation}
\delta_t = R_{t+1} + \gamma \left( \sigma Q_B(S_{t+1}, A_{t+1}) + (1 - \sigma) \sum_a \pi(a|S_{t+1}) Q_B(S_{t+1}, a) \right) - Q_A(S_t, A_t)
\label{equ:double_qsigma}
\end{equation}

and 

\begin{equation}
\delta_t = R_{t+1} + \gamma \left( \sigma Q_A(S_{t+1}, A_{t+1}) + (1 - \sigma) \sum_a \pi(a|S_{t+1}) Q_A(S_{t+1}, a) \right) - Q_B(S_t, A_t)
\label{equ:double_qsigma2}
\end{equation}

if $Q_B$ is selected. The target policy $\pi$ is computed with respect to the value function which is updated, i.e. with respect to $Q_A$ in Equation \ref{equ:double_qsigma} and with respect to $Q_B$ in Equation \ref{equ:double_qsigma2}.

Pseudocode for Double Q($\sigma$) is given in Algorithm 2. 

\begin{pseudocode}
\begin{itemize}
  \item[] Initialize $Q_A(s, a)$ and $Q_B(s, a) \quad \forall s \in \mathcal{S}, a \in \mathcal{A}$ 
  \item[] Repeat for each episode:
  \begin{itemize}
  \item[] Initialize $S_0 \neq$ terminal
  \item[] Choose $A_0$, e.g. $\epsilon$-greedy from $Q_A(S_0, .) + Q_B(S_0, .)$
  \item[] Loop for each step of episode:
  \begin{itemize}
    \item[] Take action $A_t$, observe reward $R_{t+1}$ and next state $S_{t+1}$
    \item[] Choose next action $A_{t+1}$, e.g. $\epsilon$-greedy from $Q_A(S_{t+1}, .) + Q_B(S_{t+1}, .)$
    \item[] Randomly update either $Q_A$:
    \begin{itemize}
    \item[] $\begin{aligned}
        \delta = R_{t+1} + \gamma &( \sigma Q_B(S_{t+1}, A_{t+1}) +  \\ &(1 - \sigma) \sum_a \pi(a|S_{t+1}) Q_B(S_{t+1}, a) ) -  Q_A(S_t, A_t)
    \end{aligned}$
    \item[] $Q_A(S_t, A_t) \leftarrow Q_A(S_t, A_t) + \alpha \, \delta $
    \end{itemize} \vspace{0.1cm}
    \item[] or update $Q_B$:
    \begin{itemize}
    \item[] $\begin{aligned}
        \delta = R_{t+1} + \gamma &( \sigma Q_A(S_{t+1}, A_{t+1}) + \\ &(1 - \sigma) \sum_a \pi(a|S_{t+1}) Q_A(S_{t+1}, a) ) - Q_B(S_t, A_t)
        \end{aligned}$
    \item[] $Q_B(S_t, A_t) \leftarrow Q_B(S_t, A_t) + \alpha \, \delta $
    \end{itemize}
    \item[] $A_t \leftarrow A_{t+1}$, $S_t \leftarrow S_{t+1}$
    \item[] If $S_t$ is terminal: Break
  \end{itemize}
\end{itemize}
\end{itemize}
\caption{Double Q($\sigma$)}
\label{algorithm2}
\end{pseudocode}

\section{Experiments}

In this section the performance of the newly proposed Q($\sigma, \lambda$) algorithm will be tested on a gridworld navigation task compared with the performance of classical TD control algorithms like Sarsa and Q-Learning as well as Q($\sigma$).

\begin{figure}
\centering
\begin{minipage}[b]{.35\textwidth}
  \includegraphics[width = \textwidth]{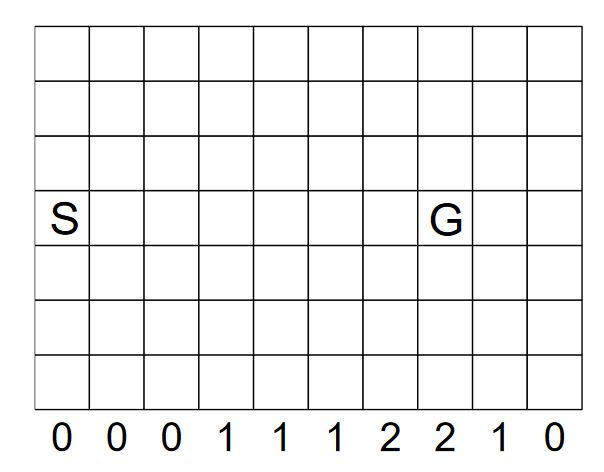}
\end{minipage} \hspace{0.5cm}
\begin{minipage}[b]{.6\textwidth}
  \includegraphics[width = \textwidth]{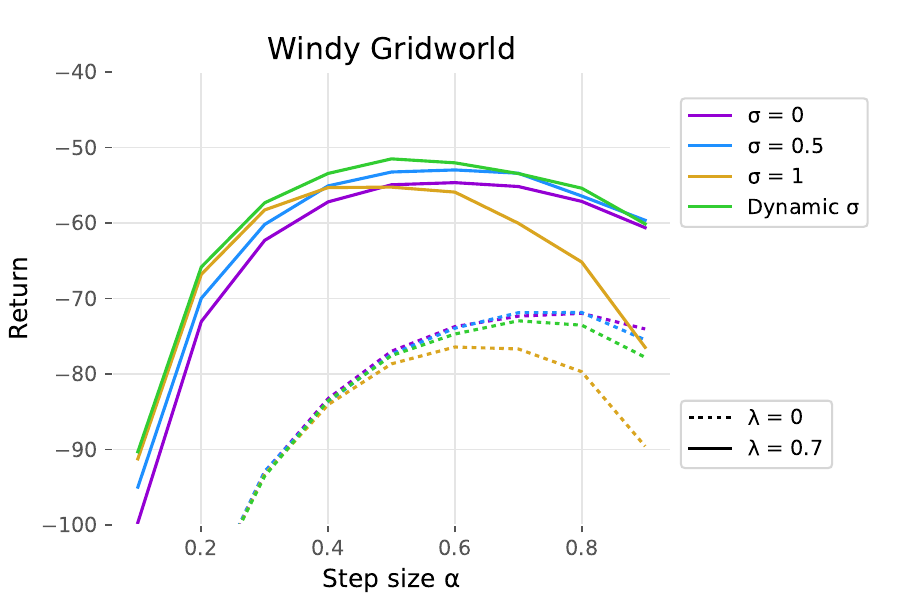}
\end{minipage}
\begin{minipage}[t]{.35\textwidth}
  \caption{The windy gridworld task. The goal is to move from the start state S to the goal state G while facing an upward wind in the middle of the grid, which is denoted in the numbers below the grid. Described by \cite{sutton1998}.}
  \label{fig:windy_gridworld}
\end{minipage}\hspace{0.5cm}
\begin{minipage}[t]{.6\textwidth}
  \caption{Stochastic windy gridworld results averaged over 100 episodes and 200 independent runs. Performance of Q($\sigma, \lambda$) for different values of $\sigma$ as a function of the step size $\alpha$. For the trace decay parameter $\lambda = [0, 0.7]$ were used. The best performance was found when using a dynamic value of $\sigma$ by multiplying $\sigma$ with a factor 0.99 after each episode.}
  \label{fig:windygrid_results}
\end{minipage}
\end{figure}

The windy gridworld is a simple navigation task described by \cite{sutton1998}. The goal is to get as fast as possible from a start state to a goal state using the actions left, right, up or down. In each column of the grid the agent is pushed upward by a wind. When an action would take the agent outside the grid, the agent is placed in the nearest cell inside the grid. The stochastic windy gridworld (\cite{deasis2017}) is a variant where state transitions are random, with a probability of 0.1 the agent will transition to one of the surrounding eight states independent of the action.
The task is treated as an undiscounted episodic task with a reward of -1 for each transition. Figure \ref{fig:windy_gridworld} visualizes the gridworld. 

Experiments were conducted using an $\epsilon$-greedy behaviour policy with $\epsilon = 0.1$. The performance in terms of the average return over the first 100 episodes was measured for different values of $\sigma$ and $\lambda$ as a function of the step size $\alpha$. For the Expected Sarsa part of the update a greedy target policy was chosen, i.e. $Q(0)$ is exactly Q-Learning. Results were averaged over 200 independent runs.

Figure \ref{fig:windygrid_results} shows that an intermediate value of $\sigma = 0.5$ performed better than Sarsa (Q(1)) and Q-Learning (Q(0)). The best performance was found by dynamically varying $\sigma$ over time, i.e. decreasing $\sigma$ by a factor of 0.99 after each episode. Multi-step bootstrapping with a trace decay parameter $\lambda = 0.7$ performed better than the one-step algorithms ($\lambda = 0$). Dynamically varying the value of $\sigma$ allows to combine the good initial performance of Sarsa with the good asymptotic performance of Expected Sarsa. This confirms the results observed by \cite{deasis2017} for n-step algorithms.

\section{Conclusions}

This paper has presented two extensions to the Q($\sigma$) algorithm, which unify Q-Learning, Expected Sarsa and Sarsa. Q($\sigma, \lambda$) extends the algorithm to an on-line multi-step algorithm using eligibility traces and Double Q($\sigma$) extends the algorithm to double learning.
Empirical results suggest that Q($\sigma, \lambda$) can outperform classic TD control algorithms like Sarsa($\lambda$), Q($\lambda$) and Q($\sigma$). Dynamically varying $\sigma$ obtains the best results.

Future research might focus on performance of Q($\sigma, \lambda$) when used with non-linear function approximation and different schemes to update $\sigma$ over time.

\bibliography{bibliography}

\end{document}